\pdfoutput=1

\documentclass[11pt]{article}

\usepackage[]{emnlp2021}

\usepackage{times}
\usepackage{latexsym}

\usepackage[T1]{fontenc}

\usepackage[utf8]{inputenc}

\usepackage{microtype}

\usepackage[ruled]{algorithm2e}
\usepackage{multirow}
\usepackage{graphicx}
\usepackage{threeparttable}
\usepackage{float}
\usepackage{amsmath}

\graphicspath{{img/}}
\title{SGPT: Semantic Graphs based Pre-training for Aspect-based Sentiment Analysis}

\author{%
Yong Qian$\dag$, Zhongqing Wang$\ddag$, Rong Xiao$\dag$, Chen Chen$\ddag$,  Haihong Tang$\dag$\\
\texttt{\{Yong Qian,Rong Xiao,Haihong Tang\}@alibaba-inc.com}\\
      {\{Zhongqing Wang,20195227080\}@suda.edu.cn}%
}

\begin{document}
\maketitle
\begin{abstract}
Previous studies show effective of pre-trained language models for sentiment analysis. 
However, most of these studies ignore the importance of sentimental information for pre-trained models.
Therefore, we fully investigate the sentimental information for pre-trained models and enhance pre-trained language models with semantic graphs for sentiment analysis.
In particular, we introduce Semantic Graphs based Pre-training(SGPT) using semantic graphs to obtain synonym knowledge for aspect-sentiment pairs and similar aspect/sentiment terms.
We then optimize the pre-trained language model with the semantic graphs.
Empirical studies on several downstream tasks show that proposed model outperforms strong pre-trained baselines. The results also show the effectiveness of proposed semantic graphs for pre-trained model. 
\end{abstract}

\section{Introduction}

Pre-trained language models learn contextualized word representations on large-scale text corpus through a self-supervised learning method, which are fine-tuned on downstream tasks and can obtain the state-of-the-art (SOTA) performance, and are pervasive and have made a tremendous impact in many NLP fields such as reading comprehension~\cite{lai2017race}, question answering~\cite{rajpurkar2016squad} and sentiment analysis~\cite{2018Deep}. 
Leveraging pre-trained language models have achieved promising results in sentiment analysis tasks, including aspect-level sentiment classification~\cite{2019A},sentence-level sentiment classification~\cite{2018Deep}. These pre-trained models have shown their power in learning general semantic representations from large-scale unlabelled corpora via well-designed pre-trained tasks. 

Sentiment analysis involves a wide range of specific tasks~\cite{liu2012sentiment}, such as sentence-level sentiment classification, aspect-level sentiment classification, aspect term extraction, and so on. 
Sentiment analysis tasks usually depend on different types of sentiment knowledge including sentiment words, word polarity, and aspect-sentiment pairs~\cite{tian2020skep}.
Recently, knowledge has been shown very important for enhancing the language representations, such as SentiLARE~\cite{ke2020sentilare} and SKEP~\cite{tian2020skep}.

Sentiment analysis especially on large-scale reviews is still very challenging, since it is hard to capture the aspect term, sentiment words from the review text.
As shown in the below example, since there are more than one aspect in the example, the traditional pre-trained model cannot capture the sentiment information. In addition, the masking language model of traditional pre-trained model will ignore the continue aspect-opinion phrase (e.g., great color).
However, based on the semantic graphs with the correlations between aspect and opinion, it is easy for proposed model to capture sentiment information, and solve the continue phrase masking problem. 

\begin{quote}
The cloth is overall good, with great \textbf{color}, but bad \textbf{material}.     
\end{quote}

To address the above challenges, we develop a semantic graph-based pre-training model to employ semantic graphs with sentiment knowledge for pre-trained model.
In particular, we explore similar aspect and sentiment words, and build a similar semantic and aspect-sentiment pair graph. 
We then employ aspect-sentiment pairs to construct the semantic graph.
Thirdly, we feed the semantic graph into a the pre-trained language model with sentimental masking.
Finally, we jointly optimize the aspect-sentiment pair prediction objective and mask language model. 
Empirical studies on several downstream tasks show that proposed model outperforms strong pre-trained baselines. The results also show the effectiveness of proposed semantic graphs for pre-trained model.
\begin{figure*}[t]
	\centering
	\includegraphics[width=320px]{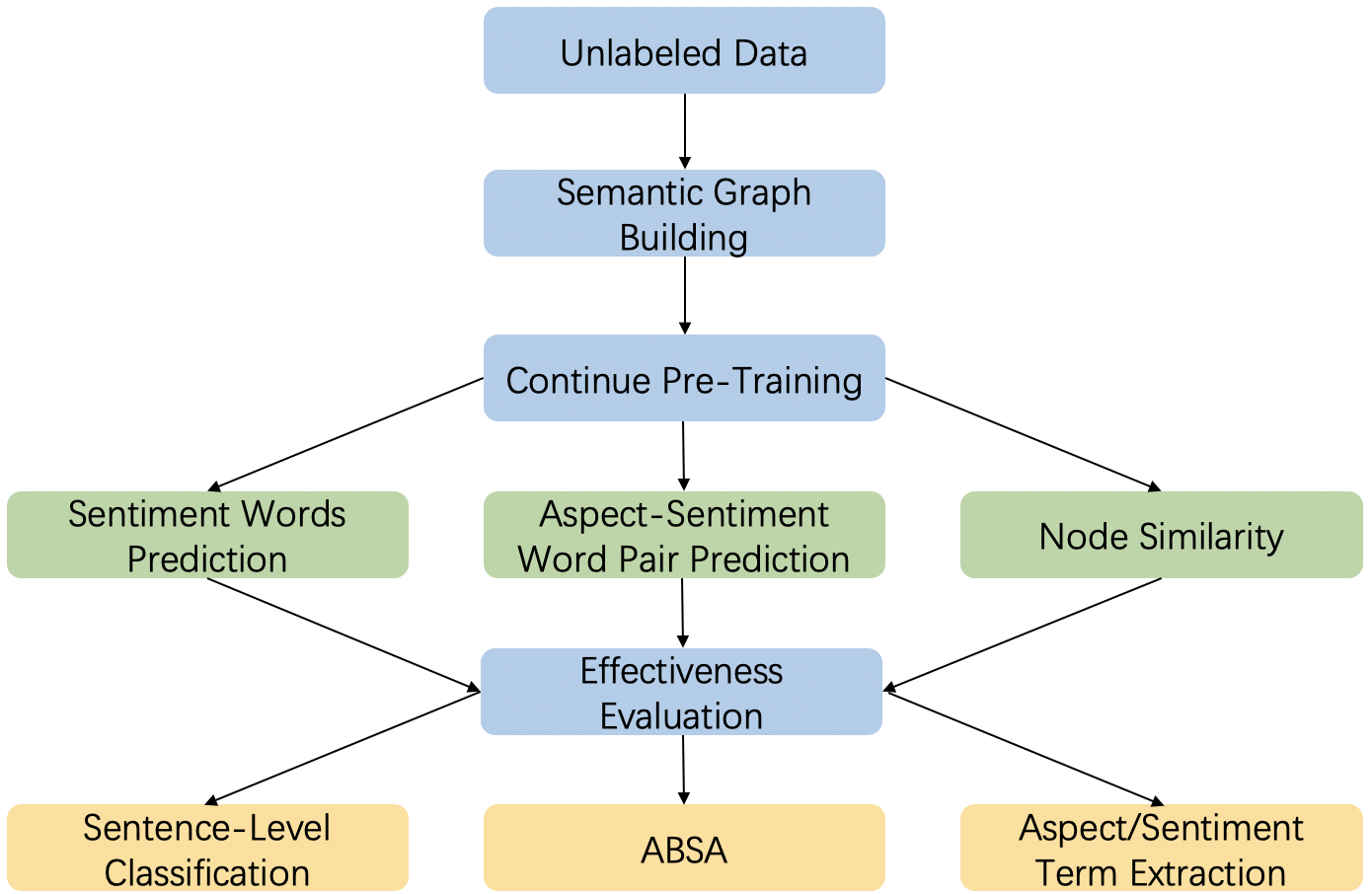}
	\caption{\label{figure-overview} The Overview of proposed models.}
\end{figure*}
In summary, our contributions are as follows:

(1) We employ semantic graphs with aspect and sentiment terms to enhance  pre-trained language models.
The results outperforms state-of-the-art models on several sentiment analysis tasks in both Chinese and English datasets.

(2) Our method significantly outperforms the strong pre-training method RoBERTa~\cite{liu2019roberta} on three typical sentiment tasks, and achieves much better results on all the datasets.

\section{Overview of Proposed Model}

In this study, we propose a semantic graph-based pre-trained model to construct the semantic graphs from aspect terms and sentiment words for the pre-trained language model.

As shown in Figure ~\ref{figure-overview}, we firstly construct a \textbf{semantic graph} with pair-wise aspect and sentiment term relations, and similarity relations among aspect and sentiment terms.
We then employ \textbf{pre-train a language model} to learn sentiment knowledge from the semantic graphs with three tasks including sentiment masking prediction, aspect-sentiment pair prediction and node similarity. 
Finally, we fine-tune the pre-training model on three \textbf{sentiment analysis tasks}: sentence-level sentiment classification, aspect-level sentiment classification and aspect/sentiment terms extraction.

\section{Semantic Graphs based Pre-training}

\begin{figure*}[t]
	\centering
	\includegraphics[width=380px]{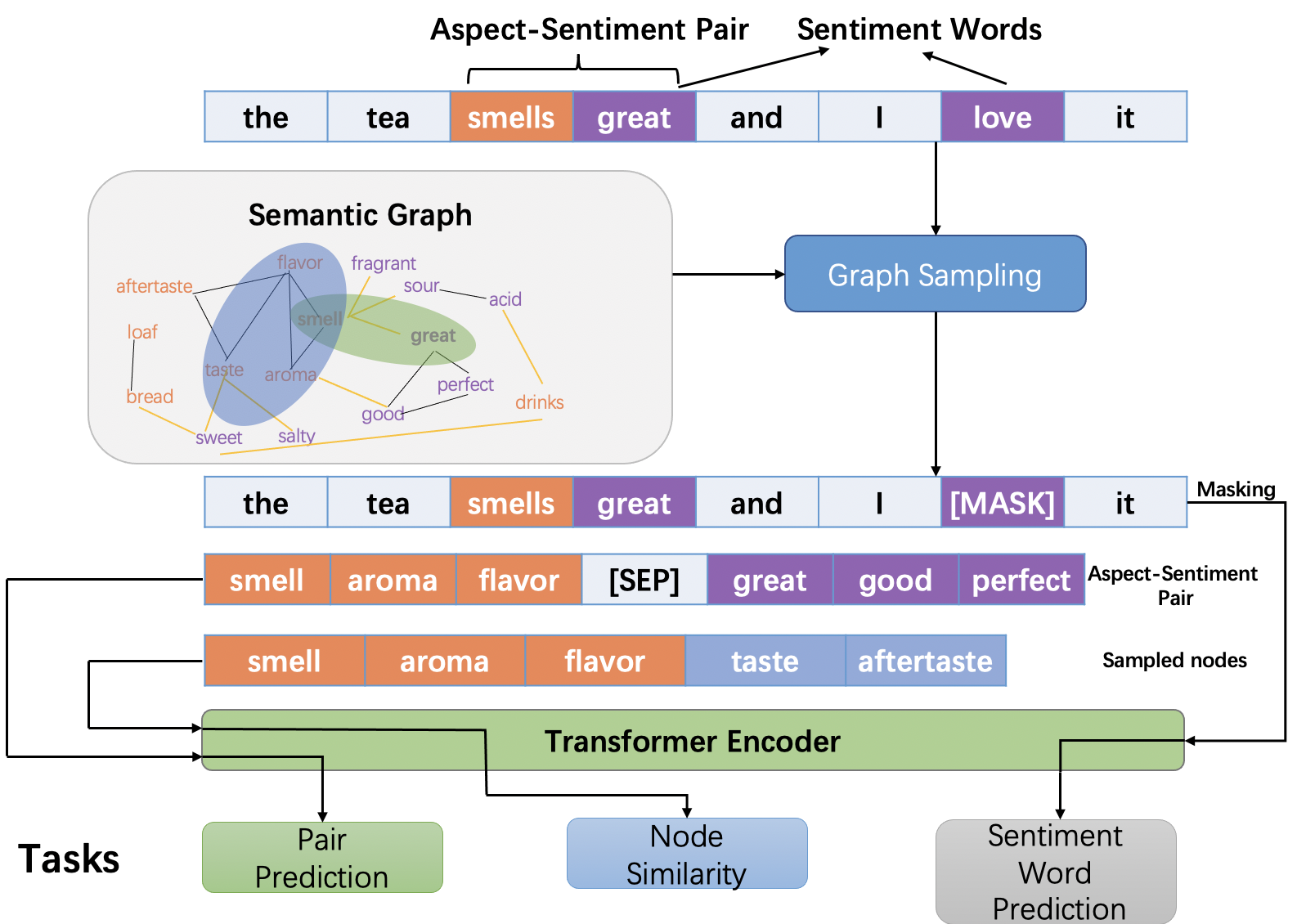}
	\caption{\label{figure-framewrok} The Framework of our pre-training paradigm. }
\end{figure*}

As shown in Figure~\ref{figure-framewrok}, we employ semantic graphs to capture sentiment knowledge for pre-trained language model.
Our paradigm contains three sentiment pre-training objectives: sentiment words masking prediction $L_{sw}$, aspect-sentiment pair prediction $L_{ap}$ and aspect-based similarity score $L_{ns}$. 

Given an input sentence, \textbf{sentimental masking prediction} attempts to recover the sentiment words masked based on the semantic graph. 
\textbf{Aspect-sentiment pairs similarity} aims to calculate the matching rate of a sentiment description on an aspect along with some related words sampled from the semantic graph. 
We extend the related words to the \textbf{aspect-based similarity score} and further learn the synonym knowledge from another perspective. 

Therefore, these three tasks are joint learning to continue pre-train the language model:
\begin{equation}
    L = L_{sw} + L_{ap} + L_{ns}
\end{equation}

\subsection{Semantic Graphs Construction}
We construct a semantic graph from large-scale unlabeled data. 
In particular, we extract the aspect words, sentiment words, and aspect-sentiment pairs from the unlabeled data. Our work is mainly based on automation methods along with a slightly manual review. 

This is a heterogeneous graph with aspect words and sentiment words as different type nodes. Two nodes are connected if they are semantic or literally similar, aspect-sentiment word pair is also connected. Our method aims to integrate those knowledge into pre-trained language model.


\subsection{Sentiment Word Prediction}
Inspired by BERT~\cite{2018BERT} that randomly replaces 15\% words with [MASK] and learn to recover them, we attempt recovering masked sentiment words to pay more attention to sentiment descriptions. For sentiment words prediction, each token in a masked sentence $\widetilde{X}$ is fed into roBERTa to get a vector representation $\mathbf{\widetilde{x_i}}$ and then normalized with softmax layer to produce a probability vector $\mathbf{\hat{y_i}}$ over the entire vocabulary. In this way, the sentiment word prediction objective $L_{sw}$ is to maximize the probability of original sentiment word $x_i$ as follows:
\begin{equation}
    \mathbf{\hat{y_i}} = \operatorname{softmax}(\mathbf{\widetilde{x_i}}W+b)
\end{equation}
\begin{equation}
    L_{sw} = -\sum_i m_i \times \mathbf{y_i}\log \mathbf{\hat{y_i}}
\end{equation}
Here, $\mathbf{W} \in R^{d \times v}$ and $\mathbf{b} \in R^{d \times 1}$ are all trainable parameters of the prediction layer. $m_i = 1$ when $x_i$ is masked, otherwise it equals 0. $\mathbf{y_i}$ is the one-hot representation of the original token $x_i$.

\subsection{Aspect-Sentiment Pair Prediction for Pre-Training}

We propose a new pre-training task to build these dependency between aspect and sentiment terms. 
Compared with predicting pairs directly, we  calculate a pair prediction over the aspect-sentiment pairs enhanced by their similar words. The  is a value between 0 to 1 which means the probability of a pair existing. For an aspect-sentiment pair, we extract a similar aspect words set $SA$ and a similar sentiment words set $SS$ from the semantic graph with algorithm 1. We concatenate $SA$ and $SS$ to construct the input sequence:
\begin{equation}
    X_{SA \oplus SS} = [CLS,SA,SEP,SS,SEP]
\end{equation}
where, CLS denotes the entire sequence representation, while SEP is a separator of two sequence. After encoding each element with roBERTa, we use $u_{cls}$ as the embedding of CLS to calculate the pair prediction. 
\begin{equation}
    \mathbf{\hat{p_i}} = \operatorname{sigmoid}(\mathbf{u_{cls}}W_p+b_p)
\end{equation}
We expect every aspect-sentiment pair equals 1. Thus, the pair prediction is:
\begin{equation}
    L_{ap} = -\sum\limits_{i=1}^{2} \mathbf{p_i}\log \mathbf{\hat{p_i}}
\end{equation}
Where $p_i$ always equals 1 when the input sequence is a pair while $p_i$ equals 0 if the input sequence is not a pair. Then the relation between aspect words and sentiment words will be established in this way.

\subsection{Node Similarity for Pre-training}
Semantic graph node similarity aim to capture model's sensitivity to aspect/sentiment synonyms. In particular, we sample synonyms from the semantic graph with similar nodes sampling algorithm detailed in Algorithm 1 and get $SA$ and $SS$. All aspect synonyms are fed into roBERTa to get a representation $U_{SA}$, so dose sentiment synonyms and gets a representation $U_{SS}$. As contrastive learning has shown great success on many areas especially unsupervised methods, we apply contrastive learning to capture these synonyms relations. The core idea of contrastive learning is to shorten the distance of positive samples and widen the distance of negative samples which perfectly meets our requirements.
\begin{equation}
\operatorname{score}\left(f(x), f\left(x^{+}\right)\right)>>\operatorname{score}\left(f(x), f\left(x^{-}\right)\right)
\end{equation}
What we want to do is clustering each similar aspect word together so all synonyms are positive samples and randomly sampled words are negative samples. We use a cosine function to measure the distance between samples and get the loss $L_{ns}$ as following:
\begin{equation}
    \operatorname{score}\left(f(x), f\left(x^{+}\right)\right) = \operatorname{cos}\left(f(x), f(x^{+}))\right)
\end{equation}

\begin{equation}
L_{ns} = -\log \frac{e^{\operatorname{score}\left(f(x), f\left(x^{+}\right)\right)}}{e^{\operatorname{score}\left(f(x), f\left(x^{+}\right)\right)}+ e^{\operatorname{score}\left(f(x), f\left(x^{-}\right)\right)}}
\end{equation}

\subsection{Sentiment Analysis with Pre-training Models}
We verify the effectiveness of our language model on three typical sentiment analysis tasks: sentence-level sentiment classification, aspect-based sentiment classification, and aspect/sentiment terms extraction. We fine-tune some strong models on the same language model as baseline to evaluate the improvement.

\subsubsection*{Sentence-level Sentiment Classification}
This task aims to classify the sentiment polarity of an input sentence. We use the final state vector of classification token [CLS] as the overall representation of an input sentence and a classification layer is added to calculate the sentiment probability on top of the transformer encoder.

\begin{algorithm}[!ht]
\caption{Similar Nodes Sampling}
\LinesNumbered 
\KwIn{Graph $G$, initial node $h$, max sampling depth $K$, max sampling words length $L$, word frequency table $T$.}
\KwOut{Sampled similar nodes $C_{h}$.}
$C_{h} \gets []$;

$S_0 \gets [h], S_1,S_2, \cdots S_K \gets []$;

\For{$k=1,2, \cdots K$}{
\For{$t \in S_{k-1}$}{
Sample all nodes linked with $t$ and append them to $S_{K}$;
}}
$\hat{C_h} = S_0 \cap S_1 \cap S_2 \cap \cdots \cap S_K$;

Sort the nodes in $\hat{C_h}$ by frequency according to table $T$ in incremental order;

Choose top-ranked nodes up to the length $L$ and append them to $C_h$;

\textbf{Return} $C_h$;
\end{algorithm}

\subsubsection*{Aspect-based Sentiment Classification}
The purpose of this task is to analyze fine-grained sentiment polarity for an aspect with a given contextual text. Thus, there are two parts in the input: contextual text and aspect description. We combine these two parts with a separator [SEP], and feed them into the language model. the final state of [CLS] also be utilized as the representation for classification.

\subsubsection*{Aspect and Sentiment Term Extraction}
This task is to extracting all aspect description or sentiment statement. The same as other tasks, all tokens are fed into language model to get a representation. Then a CRF layer is added on each tokens to predict if it belongs to an aspect or sentiment term.

\begin{table}[!ht]
  \begin{center}
    \begin{tabular}{c|c|c|c|c}
    \hline
      \multicolumn{2}{c|}{\textbf{Domain}} & \textbf{Train} & \textbf{Valid} & \textbf{Test}\\
      \hline
      \multirow{2}{*}{Furn} & POS & 37773 & 5395 & 10791 \\
      & NEG & 1795 & 256 & 512\\
      \hline
      \multirow{2}{*}{Kith} & POS & 25795 & 3712 & 7613\\
      & NEG & 2496 & 358 & 358\\
      \hline

    \end{tabular}

        \caption{\centering  Statistics of Chinese ABSA Evaluation Datasets. POS and NEG refer to positive polarity and negative polarity. There is an obviously unbalanced phenomenon, more than 10:1, between the positive and negative samples.}
            \label{table-dataset-ch} \centering 
  \end{center}
  
\end{table}

\begin{table*}[t]
\begin{center}
    \begin{tabular}{c|c|c|c|c|c|c}
    \hline
    \multirow{2}{*}{\textbf{Domain}} & \multicolumn{3}{c|}{\textbf{Extraction}} & \multicolumn{3}{c}{\textbf{Classification}} \\
    \cline{2-7}
    & \textbf{Train} & \textbf{Valid} & \textbf{Test} & \textbf{Train} & \textbf{Valid} & \textbf{Test}\\
    \hline
    Furn & 7106 & 1014 & 2029 & 120295 & 17184 & 34369\\
    \hline
    Kith & 6738 & 962 & 1925 & 92124 & 13160 & 26321\\

    \hline
    \end{tabular}
    \caption{\centering Statistics of Chinese Evaluation Datasets on Extraction and Sentence-Level Classification. The extraction task includes aspect words and sentiment words which are aggregate.}
    \label{table-dataset-ch2} \centering 
\end{center}
\end{table*}

\begin{table}[ht]
\begin{center}

    \begin{tabular}{c|c|c|c}
    \hline
    \textbf{Dataset} & \textbf{Train} & \textbf{Dev} & \textbf{Test}\\
    \hline
    SST-2 & 67K & 872 & 1821\\
    Amazon-2 & 3.2M & 400K & 400K\\
    Sem-R & 3608 & - & 1120\\
    Sem-L & 2328 & - & 638\\
    MPQA2.0 & 287 & 100 & 95\\
    \hline
    \end{tabular}
    \caption{\centering Statistics of English Evaluation Datasets. Sem-R and Sem-L refer to restaurant and laptop parts of SemEval 2014 Task 4.}
    \centering \label{table-dataset-en}
\end{center}
\end{table}

\section{Experimentation}
In this section, we introduce our training/evaluation datasets and some experiment setting. Then release the experimental results conducted from different perspectives and analyze the effectiveness compared with different baseline models.

\subsection{Data Collections}
We mainly develop with Chinese datasets and also pre-train an English version to evaluate the effectiveness on public datasets. 

The Chinese data comes from product reviews on TaoBao.com which is one of the top online shopping platforms. The Chinese pre-training dataset includes over 167 million sentences and evaluate them on two domains, i.e., Furniture (Furn) and Kitchen (Kith). 
 And all of the three tasks for Chinese are evaluated as Macro-F1 score. 
The statistics of Chinese evaluation datasets is shown in Table ~\ref{table-dataset-ch} and Table ~\ref{table-dataset-ch2}. We split all data as 7:1:2 and get the train/valid/test datasets.

The English dataset for pre-training is amazon-2 \cite{amazon2}, 3.2 million of the original training data are reserved for development. 
We evaluate the performance of the English model on a variety of English sentiment analysis datasets. Table ~\ref{table-dataset-en} summarizes the statistics of English datasets used in the evaluations.
Different tasks are evaluated on different datasets: (1) For sentence-level sentiment classification, Standford Sentiment Treebank (SST-2) ~\cite{sst}
and Amazon-2 ~\cite{amazon2} are used. The performance is evaluated in terms of accuracy. (2) Aspect based sentiment classification is evaluated on Semantic Eval 2014 Task4 \cite{semeval14}. This task contains both restaurant domain and laptop domain, whose accuracy is evaluated separately. (3) For the extraction task, MPQA 2.0 ~\cite{mpqa1, mpqa2} dataset is used which aims to extract the aspects or the holders of the sentiments. We measured with the method in SRL4ORL ~\cite{SRL4ORL}, which is released and available online.

\begin{table}[t]
    \begin{center}
    \begin{tabular}{c|c}
    \hline
    \textbf{Parameter} & \textbf{Value}\\
    \hline
    Dataset Size CN & 167M\\
    Dataset Size EN & 3.2M\\
    Masking Rate & 20\%\\
    Number of Pairs & 2\\
    Sentence Length & 512\\
    Learning Rate & 1e-5\\
    Batch Size & 32\\
    Warm up Ratio & 0.1\\
    Weight Decay & 0 \\
    \hline
    \end{tabular}
    \caption{\centering Pre-Training Parameters.}
    \label{table-paraeter}
    \end{center}
\end{table}

\begin{table*}[ht]
\begin{center}

    \begin{tabular}{c|c|c|c|c|c|c|c|c}
    \hline
    \multirow{3}{*}{\textbf{Model}} &
    \multicolumn{2}{c|}{\textbf{Aspect-Level}}&
    \multicolumn{2}{c|}{\textbf{Sentence-Level}}& 
    \multicolumn{4}{c}{\textbf{Extraction}}  \\
    \cline{2-9}
    & \textbf{Furn}& \textbf{Kith}
    &  \textbf{Furn}& \textbf{Kith}
    & \multicolumn{2}{c|}{\textbf{Furn}}
    & \multicolumn{2}{c}{\textbf{Kith}}\\
    \cline{2-9}
    & M-F1 & M-F1 & M-F1 & M-F1 & Aspect & Sentiment& Aspect & Sentiment\\
    \hline
    BERT &90.9 & 91.5& 90.2  & 93.5&79.9&82.0&77.2&79.2  \\
    RoBERTa &94.1 & 94.3& 92.8 & 95.3&81.4&83.1&78.6&81.6  \\
    SKEP &94.6 & 94.9&  92.2 & 95.5&82.7&84.1&80.3&83.3  \\
    \hline
    Pair Prediction & 95.2 & 95.8& 92.6 & 94.1&84.2&85.2&82.8&84.8  \\
    Node Similarity & 95.4 & 95.9& 93.2&96.4&84.8&86.9&83.0&84.7 \\
    Ours & \textbf{96.2} & \textbf{96.7}& \textbf{93.2} & \textbf{96.8}&\textbf{85.0}&\textbf{87.3}&\textbf{83.8}&\textbf{85.0} \\
    \hline
    \end{tabular}
    \caption{\centering Results of Chinese Evaluation on Extraction and Sentence-Level Classification. Model BERT and RoBERTa refer to fine-tune directly on downstream tasks. Pair prediction refer to our pre-training methods with sentiment masking and pair prediction. Node similarity means sentiment masking with node similarity. Ours means the complete method we propose. M-F1 is the abbreviation of macro-F1 score.}
    \label{table-result-ch} \centering 
\end{center}
\end{table*}

\begin{table*}[!tp]
\begin{center}
    \begin{tabular}{c|c|c|c|c|c|c}
    \hline
    \multirow{2}{*}{\textbf{Model}} &
    \multicolumn{2}{c|}{\textbf{Aspect-Level}}& 
    \multicolumn{2}{c|}{\textbf{Sentence-Level}} & \multicolumn{2}{c}{\textbf{Extraction}}\\
    \cline{2-7}
     & Sem-L & Sem-R & SST-2 & Amazon-2 & MPQA-Holder & MPQA-Target\\
    \hline
    Previous SOTA & 81.35 & 87.89& 97.10 & 97.37  & 83.67/77.12 & 81.59/73.16\\
    SKEP & 81.32 & 87.92& 96.70 & 96.94  & 84.25/79.03 & 82.77/74.82\\
    \hline
    Pair Prediction & 82.46 & 88.79& 97.18 & 98.02  & 85.31/80.89 & 83.48/76.43\\
    Node Similarity & 84.31 & \textbf{89.04}& 98.15 & 98.25  & \textbf{86.42}/81.78 & 84.69/77.01\\
    Ours & \textbf{84.45} & 89.03& \textbf{98.27} & \textbf{98.33}  & 86.16/\textbf{82.05} & \textbf{84.79}/\textbf{77.56}\\
    \hline
    \end{tabular}
    \caption{\centering Results of English Evaluation}
    \label{table-result-en} \centering 
\end{center}
\end{table*}

\subsection{Data Pre-Processing}
To build the semantic graph, we extract these information as following: 

\textbf{Aspect/Sentiment Term Extraction} Aspect and Sentiment descriptions are extracted by BERT-CRF which is trained on labeled datasets up to 85 F1 score.

\textbf{Aspect-Sentiment Pair Extraction} We match the aspect-sentiment pairs with simple constraints. An aspect-sentiment pair refers to the mention of an aspect and
corresponding sentiment words. Thus, a sentiment word with its nearest aspect has high probability to be a pair. More specifically, we limit the aspect-sentiment pair must be included in one sentence and only one-to-one pairs are considered.

\textbf{Similar Words Extraction} As similar words can be seen as words with a same category, we employ DBSCAN clustering algorithm to get coarse-grained synonyms represented by the average pooling of all piece words' Word2Vec embeddings. A recycling mechanism is applied to further clustering big clusters, which contain lots of irrelevant words, into small clusters by grid searching different parameters. Finally, we review all similar words manually to get more accurate synonyms.


\subsection{Experiment Setting}
We use RoBERTa as the base language model and continue pre-training it with our paradigm. We concatenate different sentences until the sentence length up to 512 and train them with batch size 32. An adam optimizer is applied with learning rate 1e-5 and warmup ratio 0.1 without weight decay. 
For sentiment masking, we mask sentiment words as far as possible up to 20\% and two aspect-sentiment pairs at most needed to be extracted. 
At the stage of fine-tune for downstream tests, we take the same parameters as SKEP ~\cite{tian2020skep} did.

\subsection{Comparison with Baselines}
We compare SGPT with two strong pre-training baselines:  RoBERTa and a similar continue pre-training method SKEP. 

The results on Chinese datasets are shown in table \ref{table-result-ch}. For classification tasks, there is no obvious improvement from RoBERTa to SKEP while SGPT outperforms SKEP 1.6 and 1.8 on aspect-level, along with 1.0 and 1.3 on sentence-level. Meanwhile both pair prediction and node similarity performs better than SKEP. More details, SGPT overcomes the unbalanced problem with average 3 point improvements on negative samples. 
For extraction task, sentiment continuing pre-training's effectiveness is also significantly. SKEP has a 1-2 point improvement compared with RoBERTa and SGPT has a further 2-3 point improvement based on SKEP.

Meanwhile, English experiments are shown in table \ref{table-dataset-en}. SGPT outperforms other methods over three task as well as in Chinese datasets and indicates that SGPT is universal to cover different language and different tasks.

\begin{figure*}[t]
	\centering
	\includegraphics[width=400px]{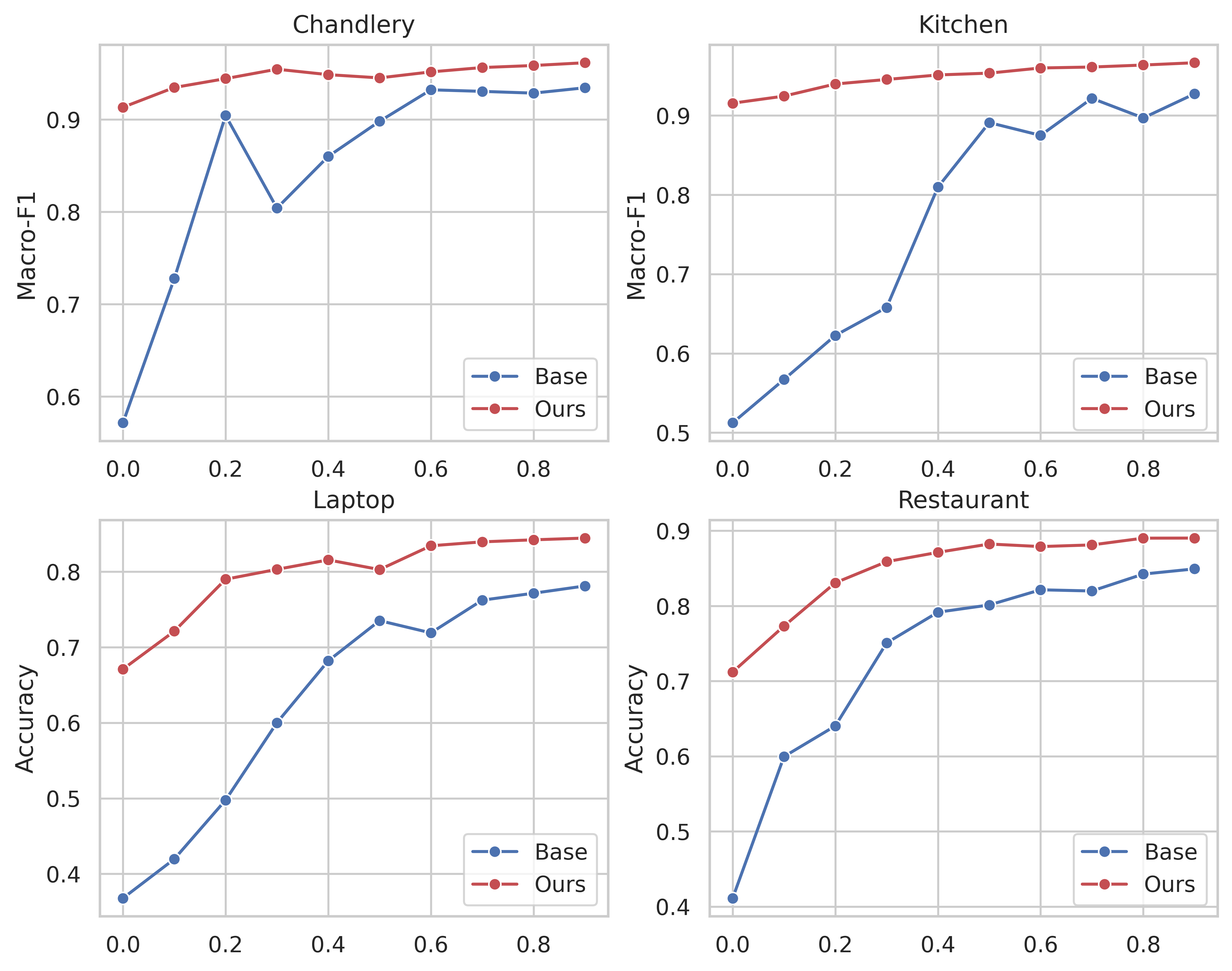}
	\caption{\label{figure-scale} The Results under Different Data Scales.}
\end{figure*}

\subsection{Influence of Different Factors}

We then analyze the influence of different factors with both pair prediction and node similarity separately to verify the influence factors on Table~\ref{table-dataset-en}.

\textbf{Effective of Aspect-Sentiment Pair Prediction}
For aspect based sentiment analysis, we propose the aspect-sentiment pair prediction to build the dependence between aspect words and sentiment words. The results reflect that continuing pre-training language model with pair prediction can absorb pairs information successfully and adapt great on downstream task related to aspect or sentiment. 

\textbf{Effective of Node Similarity}
The node similarity aims to learn the synonym knowledge from semantic graphs. Compared with aspect-sentiment pair prediction, the experiment results shows that it almost outperforms far more than all pair prediction's results except aspect term extraction. We think node similarity is an even more powerful mechanism for gaining sentiment information and contrastive learning also plays a great role in this.

Finally, we combine the two factors into one model. The results shows that our composite method benefits from the two with different advantages and get a better performance than both of them separately. 
\subsection{Influence of Training Size}
To verify how capable SGPT solves the unbalance label problem and few-shoot problems, we design an experiment which increases the data scale from 10\% to 100\%, then compare the effectiveness between SGPT and fine-tuning directly. As shown in figure ~\ref{figure-scale}, SGPT has great performance even on 10\% of all training data and gets more than 30 point improvement compared with the base model. The base model is close to SGPT until 60\% of data has been used but still has a big gap with SGPT.

\begin{table*}[!ht]
    \centering
    \begin{tabular}{c|c|c|c|c|c}
    \hline
    \textbf{ID} & \textbf{Review} & Sentiment Polarity &RoBERTa & SKEP & Ours\\
    \hline
    1 & \begin{tabular}[c]{@{}l@{}} This air conditioner costs \textbf{less} \emph{electricity} \\and it's refrigeration is very good \end{tabular} & POS& NEG & NEG & POS\\
    \hline
    2 & \begin{tabular}[c]{@{}l@{}}Perfect! Although the \emph{price} is \textbf{high},\\ the quality is very good. \end{tabular} & NEG & POS & POS & NEG\\
    \hline
    
    \end{tabular}
    \caption{Case Study. Italics words are aspect terms and bold words are sentiment descriptions.}
    \label{case}
\end{table*}
\subsection{Case Study}
We give two aspect based sentiment analysis examples  in table ~\ref{case} and illustrate the situation where RoBERT or SKEP can't solve but can be overcame by SGPT. 

The first example is about unconventional expression. For most of reviews, "cost" usually means needing more and SKEP makes the decision from this inertia thinking. While SGPT takes more words into consideration that linked in the semantic graph and recognizes "costs less" is a positive expression.

The second example is fine-grain sentiment polarity contrary to the sentence-level polarity. The pair prediction task in SKEP can't identify every aspect-sentiment pair when the sentiment description is intensive so that it assesses "price" from a holistic perspective. While SGPT benefits from pair prediction without predicting all pairs and recognizes the real sentiment polarity.

\section{Related Works}

\subsection{Sentiment Analysis with Knowledge}  

Various types of sentiment knowledge, including sentiment words, aspect-sentiment pairs and prior sentiment polarity from Senti WordNet ~\cite{ke2020sentilare},  have been proved to be useful for a wide range of sentiment analysis tasks. Sentiment words with their polarity are widely used for sentiment analysis, including sentence level sentiment classification ~\cite{2011Lexicon,2017Lexicon,2018Deep}, aspect-level sentiment classification ~\cite{vo2015target,2019A}, sentiment extraction ~\cite{li2017deep}, emotion analysis ~\cite{gui2017question,fan2019knowledge} and so on. Lexicon-based method ~\cite{turney2002thumbs,taboada-etal-2011-lexicon}directly utilizes polarity of sentiment words for classification. Traditional feature-based approaches encode sentiment word information in manually-designed features to improve the supervised models ~\cite{bakshi2016opinion,agarwal2011sentiment}. In contrast, deep learning approaches enhance the embedding representation with the help of sentiment words ~\cite{shin2016lexicon}, or absorb the sentiment knowledge through linguistic regularization ~\cite{qian2016linguistically}.Aspect-sentiment pair knowledge is also useful for aspect-level classification and sentiment extraction. Previous works often provide weak supervision by this type of knowledge, either for aspect level classification~\cite{zeng2019variational} or for sentiment extraction ~\cite{yang2017transfer,ding2017recurrent}. 
One related study is SKEP~\cite{tian2020skep}, which utilize sentiment knowledge to embed sentiment information at the word, polarity and aspect level into pre-trained sentiment representation. But it's hard to recover aspect-sentiment pair due to aspect-sentiment pair appear continuously(85$\%$ probability) in product review(product reviews in tabao.com ).

\subsection{Knowledge Enhanced PLMs} 
Recently, to enable PLMs with world knowledge, several attempts ~\cite{wang2019kepler,peters2019knowledge,zhang2019ernie,liu2020k,wang2020k} have been made to inject knowledge into BERT leveraging Knowledge Graphs (KGs). Most of these work adopts the "BERT+entity linking" paradigm, whereas, it is not suitable for E-commerce product reviews due to the lack of quality entity linkers as well as KGs in this domain. Skep conducts aspect-sentiment pair masking, sentiment word masking,common token masking and utilize three sentiment knowledge prediction objectives, with sentiment word prediction, word polarity prediction and aspect-sentiment pair prediction and aspect-sentiment pairs is converted into multi-label classification. However, aspect-sentiment word appear continuously in product reviews, which make it difficult to predict when masking the pair. Our work also differs from the work skep, we develop a novel pre-training paradigm that leverage semantic graphs and incorporate sentiment knowledge into pretraining, a detailed comparison between our model and skep pre-trained language models can be found in $\S 3$.

\section{Conclusion}
In this study, we design three pre-training tasks to continue pre-training language model for downstream sentiment analysis tasks. We propose a semi automatic method to build a semantic graph and employee language model to adopt the graph knowledge with these tasks. Sentiment words masking for paying more attention to sentiment term, aspect-sentiment pair prediction for building the dependence between aspect and sentiment, node similarity for learning synonym knowledge, which all get great improvement over different downstream tasks and different languages.

In the future, we will try to apply SGPT on more
sentiment analysis tasks, to further see the generalization of SGPT, and we are also interested in exploiting more efficiency method of building semantic knowledge.

\bibliography{anthology,custom}
\bibliographystyle{acl_natbib}

\end{document}